%% file: root.tex
\documentclass[lettersize,journal]{IEEEtran}
\input{macros}
\begin{document}

\title{Evaluating and Improving the Robustness of LiDAR Odometry and Localization Under Real-World Corruptions}

\author{Bo Yang,
    Tri Minh Triet Pham,
    and~Jinqiu Yang%
    \thanks{Bo Yang, Tri Minh Triet Pham, and Jinqiu Yang are with the Department of Computer Science and Software Engineering, Concordia University, Montreal, Quebec, Canada (e-mail: \{b\_yang20, p\_triet, and jinqiuy\}@encs.concordia.ca).}% <-this % stops a space
}

\markboth{Journal of \LaTeX\ Class Files,~Vol.~14, No.~8, August~2021}%
{Shell \MakeLowercase{\textit{et al.}}: A Sample Article Using IEEEtran.cls for IEEE Journals}


\maketitle

\begin{abstract}
LiDAR odometry and localization are two widely used and fundamental applications in robotic and autonomous driving systems. 
Although state‑of‑the‑art (SOTA) systems achieve high accuracy on clean point clouds, their robustness to corrupted data remains largely unexplored.
We present the first comprehensive benchmark to evaluate the robustness of LiDAR pose‑estimation techniques under 18 realistic synthetic corruptions.
Our results show that, under these corruptions, odometry position errors escalate from 0.5\% to more than 80\%, while localization performance stays consistently high.
To address this sensitivity, we propose two complementary strategies. 
First, we design a lightweight detection-and-filter pipeline that classifies the point cloud corruption and applies a corresponding filter (e.g., bilateral filter for noise) to restore the point cloud quality. 
Our classifier accurately identifies each corruption type, and the filter effectively restores odometry accuracy to near-clean data levels. 
Second, for learning‑based systems, we show that fine-tuning using the corrupted data substantially improves robustness across all tested corruptions and even boosts performance on clean point clouds on one data sequence. 
\end{abstract}


\section{Introduction}
%Autonomous robots and vehicles rely on real-time pose estimation to determine their position and orientation, enabling navigation, collision avoidance, mapping, and path planning.
%Due to LiDAR enables precise environmental perception by generating high-resolution 3D point cloud data (PCD) independent of external lighting conditions. 
%As a result, it is widely used in pose estimation, particularly in scenarios where .
%LiDAR pose estimation is realized by two concrete applications,  i.e., odometry~\cite{odometry_survey} and localization~\cite{yin2024survey}.
%In particular, LiDAR odometry estimates an agent's incremental motion between consecutive LiDAR scans to enable reliable self-tracking and map building in unknown environments. LiDAR localization, by contrast, computes an agent's absolute pose within an existing map, supporting accurate navigation in structured settings such as urban road networks.
% \bo{TODO: 1. double check all places mentioning of the full results and code available in appendix or public artifact;}
Reliable pose estimation is critical in robotics and autonomous driving systems (ADS), where even minor errors can lead to catastrophic failures in navigation, planning, and control~\cite{Bukhori2017DetectionOK, driftwithdevil}. 
LiDAR-based pose estimation is widely adopted for its superior reliability and performance, particularly when Global Navigation Satellite System (GNSS) signals are unreliable (e.g., urban canyons) or unavailable (e.g., indoors). 

To this end, extensive research has focused on improving LiDAR pose estimation, i.e., odometry and localization.
Particularly, LiDAR odometry estimates an agent’s incremental motion between consecutive LiDAR scans, enabling reliable self-tracking and map building in unknown environments. In contrast, LiDAR localization determines an agent’s absolute pose within a pre-built map, supporting accurate navigation in structured settings such as urban road networks.
These methods vary in their registration strategies, including direct scan alignment \cite{ndt, Vizzo2022KISSICPID}, feature-based matching \cite{zhang2014loam}, and, more recently, learning-based approaches \cite{deepvcp, 9561063} to improve generalization across diverse environments. 
%\todo{Start with ``To this end, many works are proposed to improve LiDAR ....'' such techniques differ ... such as xxx yyy zz; then mention how the errors aggregated ; and localization impact the critical designs of robotic and ads.} 
%LiDAR odometry estimates incremental motion by registering consecutive scans, making it essential for exploring unknown environments. 
%However, odometry suffers from drift due to error accumulation over time. 
%On the other hand, LiDAR localization aligns online scans with pre-built maps to achieve drift-free global pose estimation, which is particularly crucial in safety-critical applications, such as fully autonomous vehicles that require global localization accuracy at the centimeter level~\cite{8461224}. 
%Despite the different objectives, both tasks share underlying point cloud registration techniques to obtain fine poses. 
%Recent advancements of deep learning in computer vision has enabled learning-based LiDAR odometry \cite{Du_Xu_Li_Qu_Fu_Liu_2025, deepvcp}, which are prominent in direct and feature matching methods.

Despite the success, LiDAR odometry and localization face inherent challenges on robustness caused by non-clean LiDAR data due to real-world corruptions. 
Prior works reveal that point cloud data (PCD) may be corrupted or noisy, caused by real-world environment challenges, such as adverse weather~\cite{ars-9-49-2011}, sensor noises~\cite{Hongchao2012AnalysisOP} or dusts~\cite{dust_noise}, reflective surfaces~\cite{10588461}, and occlusions caused by dynamic environments~\cite{laconte2023toward}. 
% Although state-of-the-art (SOTA) LiDAR odometry and localization methods achieve high accuracy under clear conditions, their performance can degrade significantly when faced with corrupted or noisy PCD~\cite{Zhang20243DLS}\todo{this citation is not correct. What does it mean? this paper studies the performrance of lidar under such scenarios already?}\bo{This paper mentions that LiDAR SLAM faces challenges in rain and fog weather}. 
This issue is particularly pronounced in learning-based approaches, which often lack transparency and interpretability, making it difficult to understand their decision-making processes and limiting their ability to generalize to scenarios that diverge from the training data \cite{Zhang20243DLS}.

While prior studies have examined specific aspects of robustness, e.g., occlusion in ~\cite{laconte2023toward} and security threats via adversarial attacks ~\cite{yoshida2022adversarial}, these works are limited in scope and do not offer a comprehensive robustness benchmark for LiDAR pose estimation systems. Notably, there is \textbf{no systematic evaluation of how a wide range of real-world point cloud corruptions affect LiDAR odometry and localization}, particularly in learning-based approaches, which are known to be more sensitive to data quality. Furthermore, existing efforts \textbf{fall short in proposing methods to improve the robustness} of LiDAR pose estimation against such corruptions. %This gap leaves the field without standardized tools or baselines for assessing or enhancing robustness—despite its critical role in safety-sensitive applications.

%The extent of the performance degradation under such corruptions remains unclear, posing serious risks which can lead to critical failures, e.g., a robot can become lost navigating a cluttered warehouse where the environment or the robot itself is frequently moved causing significant changes to consecutive PCD frames or an autonomous vehicle driving off-road due to misinterpreted boundaries and obstacles cause by corrupted PCD.

In this work, we propose a systematic evaluation framework, \textbf{\textit{\tool}}, to address the research gap regarding the impact of PCD corruptions on LiDAR odometry and localization, especially learning-based methods which are known to be sensitive to such corruptions.
Furthermore, to mitigate these adverse effects, we further introduce and experiment with two complementary strategies to enhance robustness.
% \todo{Bo, the current results shows that detection-and-filter enhance robustness for both right?}\bo{Yes}\triet{YOu should fix the following 2 sentences, same for their equivalent in the abstract}
First, we design \textbf{a novel and lightweight detection‑and‑filter pipeline} and validate its effectiveness on noise‑induced corruptions. 
Second, for learning‑based methods, we show that \textbf{fine-tuning on corrupted data significantly enhances robustness}.
Our key contributions are as follows:
\begin{itemize}
    % \item \todo{We propose [name], a novel comprehensive robustness benchmark for LiDAR odometry/localization systems.} We systematically evaluate five SOTA LiDAR odometry/localization systems against 18 realistic point cloud corruptions which uncovers key vulnerabilities for each representative system, aiding the development of robust and reliable \todo{xxx} systems.
    \item We propose \textbf{\textit{\tool}}, a novel comprehensive robustness benchmark for LiDAR odometry/localization systems.
    % For instance, ICP fails when background noise breaks scan correspondences and feature-based pipelines degrade sharply under density drops.  
    \item We propose a novel lightweight detection-and-filter method which enables accurate, efficient, generalizable classification and restoration of corrupted PCD. 
    % \item We propose an accurate, lightweight corruption classification model which enables generalizable, efficient, and accurate noise detection and classification.
    \item We investigate various mitigation strategies, including detection-and-filter and fine-tuning to defend the LiDAR odometry/localization systems against PCD corruptions. 
    \item Public release of code and data for reproducibility\footnote{\url{https://doi.org/10.5281/zenodo.17128983}}.
\end{itemize}

Our framework reveals that SOTA LiDAR odometry is sensitive to at least one type of PCD corruptions. 
Our detection-and-filter can effectively defend the noise-induced corruptions for non-learning systems and corruption augmented fine-tuning can bring learning-based systems to near-clean level. 
% \todo{Our contribution [mention significance and difference from previous works.]}

\section{Background and Related Works}

\subsection{LiDAR-Based Odometry and Localization}
% \todo{this is not really backgrond; background should talk about the setting of pose estimation, which should start with 'first LiDAR scan', then  `real-time scan'}
LiDAR-based odometry and localization allow autonomous agents to track their motions (odometry) and get their absolute position and orientation in a map (localization) in 3D environments. 
The core of LiDAR-based odometry and localization is point cloud registration, 
% i.e., aligning LiDAR scans to estimate their 6-degree-of-freedom (DoF) spatial transformations (translation and rotation). \todo{improve transition}
% Point cloud registration (or scan matching) is 
the process of aligning two LiDAR scans to determine their relative poses. 
% This technique is fundamental to LiDAR localization, which is responsible for data association and initial pose estimation of adjacent LiDAR scans.

\subsubsection{Point Cloud Registration}\label{sec:registration}
% \jinqiu{this is very dry to read; there is no insight; you need to organize this for better presetantion; this is not a survey paper...}
Given a source point cloud $\mathcal{P} \in \mathbb{R}^{3 \times N}$ and a target point cloud $\mathcal{T} \in \mathbb{R}^{3 \times M}$, point cloud registration solves:
\begin{equation}
    \mathbf{R}^*, \mathbf{t}^* = \underset{\mathbf{R} \in SO(3), \mathbf{t\in \mathbb{R}^3}}{\arg\min} \ \text{dist} \left(\mathbf{R} \mathcal{P} + \mathbf{t}, \mathcal{T}\right)
\end{equation}
where $\mathbf{R}$ is a 3-DoF rotation matrix and $\mathbf{t}$ is a 3-DoF translation vector. The source point cloud is rotated by $\mathbf{R}$ and translated by $\mathbf{t}$.
% The operation $\mathbf{R} \mathcal{P} + \mathbf{t}$ rotates all points using $\mathbf{R}$ and translates them by $\mathbf{t}$. 
The function $\text{dist}(\cdot, \cdot)$ quantifies the errors between the corresponding points in $\mathbf{R} \mathcal{P} + \mathbf{t}$ and $\mathcal{T}$. 
In other words, this equation seeks the optimal pose $\mathbf{R}^*, \mathbf{t}^*$ that aligns the source point cloud $\mathcal{P}$ to the target point cloud $\mathcal{T}$. 

Various approaches are proposed to solve this problem, evolving from geometry-based matching methods to modern learning-based methods. 
They can be broadly categorized into direct matching, feature-based, projection-based, and neural distance field-based methods. 
Direct methods like iterative closest point (ICP)~\cite{icp, Vizzo2022KISSICPID} align raw points by iteratively minimizing the distance between nearest neighbor points. 
% Vanilla ICP struggles with problems such as sensitivity to initialization and noises. Variants of ICP have been developed to solve these problems and ICP-based methods remain competitive and widely adopted~\cite{Vizzo2022KISSICPID}. 
Normal distribution transform (NDT) is another classic direct method~\cite{ndt}, which partitions the target point cloud into a grid of cells, with each cell modeling the local distribution of points as a Gaussian. 
Instead of minimizing the distances between points, it optimizes the likelihood measure that evaluates how well the source point cloud fits the normal distribution in the target grid. 
A recent breakthrough is LiDAR Odometry and Mapping (LOAM)~\cite{zhang2014loam} which advances the point cloud registration by extracting robust features (e.g., edges and planar regions) from point clouds and matching the features. 
With the development of deep learning, researchers further improved feature extraction from hand-crafted features to learned features~\cite{deepvcp, 9561063}. 
% Another approach leveraging deep learning projects 3D point clouds into 2D images and leverages convolutional neural networks to learn the features~\cite{9561063}.   
The success of Neural Radiance Fields (NeRF)~\cite{mildenhall2020nerf} in modeling implicit 3D scenes has inspired the efforts to explore of the application in the LiDAR-based pose estimation~\cite{wiesmann2023ral, Deng_2023_ICCV}.

\subsubsection{LiDAR Odometry}
LiDAR odometry tracks real-time motion by aligning consecutive scans. While point cloud registration offers high accuracy, accumulated errors lead to significant long-term drift. To correct this, post-processing techniques like loop closure~\cite{newman2005slam} are often used. SOTA odometry systems cover the various approaches as discussed above~\cite{Vizzo2022KISSICPID,zhang2014loam,9561364,9561063,deepvcp,Deng_2023_ICCV}.
% LiDAR odometry tracks the accurate motion of autonomous agents in real-time by aligning consecutive LiDAR scans. 
% Despite the high accuracy of point cloud registration, the long-term trajectory drifts significantly due to accumulated errors. 
% To mitigate the drift, additional algorithms such as loop closure \cite{newman2005slam} should be applied in post-processing.
% SOTA odometry systems cover all the approaches mentioned in the previous section \cite{Vizzo2022KISSICPID,zhang2014loam,9561364,9561063,deepvcp,Deng_2023_ICCV}. 

\subsubsection{LiDAR Localization}
LiDAR localization determines the global 6-DoF pose by aligning an online scan with a pre-built map, enabling centimeter-level accuracy critical for fully autonomous driving systems.
While some localization systems estimate only 3-DoF position, our work focus on 6-DoF pose estimation.
Compared to scan-to-scan registration in odometry, scan-to-map registration in localization is more challenging due to the vast search space. 
To address this problem, most methods first obtain an initial pose via external signals (e.g., GNSS)~\cite{8461224} or global position recognition~\cite{yin2024survey}, then perform the point cloud registration to estimate the fine pose. 
Although one-stage global pose estimation approaches exist, they generally underperform compared to those with initialization~\cite{yin2024survey}. 
Similar to odometry, SOTA localization approaches also cover the full spectrum of registration methods~\cite{8461224,6856596,L3NET_2019_CVPR,wiesmann2023ral}.

\subsubsection{Multi-Sensor Fusion Approaches}
Multi-sensor fusion (MSF) combines complementary sensors, such as LiDAR, cameras (vision odometry and localization), IMUs (high-frequency motion data), and GNSS (global localization), to enhance pose estimation~\cite{8461224}.
In this work, we focus on LiDAR-only pose estimation systems to study LiDAR's standalone sensitivity to noises in pose estimation. 

\subsection{Robustness of LiDAR-based Systems}
Previous research has extensively explored the robustness and security of LiDAR-based 3D obstacle detection. Several studies have also introduced realistic adverse weather simulations to evaluate the impact of environmental conditions on PCD, including rain~\cite{kilic2021lidar,hegde2023source}, snow~\cite{HahnerCVPR22}, and fog~\cite{HahnerICCV21}.
% Sun2022BenchmarkingRO
% This includes benchmarking against perturbations~\cite{yu2022benchmarking, Gao2023BenchmarkingRO, Albreiki2022OnTR, 10286105, dong2023benchmarking}; 
% testing and adversarial attacks via point-based~\cite{metamorphic_lidar, ZHENG2023109825, Xu2021AdversarialAA, lidar_occulsion, adversarial_locations, Zhu2021AdversarialAA, gan_attack, liu_multiview, WANG202127} and object spoofing methods~\cite{adversarial_lidar_attack, Illusion_and_Dazzle, invisibleobject, Abdelfattah2021, Yang2021RobustRP, pmlr-v164-tu22a, Tu_2020_CVPR}. 
Researchers have also systematically benchmarked 3D obstacle detection under common point cloud corruptions~\cite{10286105} and investigated its vulnerability to adversarial attacks~\cite{adversarial_lidar_attack, Liao_Yan_Zhang_Zhai_Wang_Fu_2025}.

For LiDAR pose estimation, recent work has investigated the security of localization and odometry against adversarial attacks~\cite{yoshida2022adversarial,zhang2024prepared,10591887}. The study most closely related to ours is Laconte et al.~\cite{laconte2023toward}, which derives a closed-form approximation of the worst-case localization error for the ICP algorithm~\cite{icp} as a function of the number of corrupted points. Their theoretical bound offers useful insight into how adversarial outliers can degrade a linearized (vanilla) ICP. However, SOTA odometry and localization systems typically employ more advanced ICP variants or alternative registration approaches, and the idealized bounds in~\cite{laconte2023toward} do not fully reflect the degradation under realistic corruptions. 
Our work complements this research gap by performing a systematic empirical evaluation of a broad set of odometry and localization pipelines, including both learning-based and classical registration methods, under a comprehensive suite of realistic corruption scenarios.

\section{Methodology}\label{sec:approach}
% In this section, we describe the two components of our proposed robustness framework for LiDAR SLAM, one to evaluate the robustness of LiDAR SLAM through emulating complex environment changes, and one to improve the robustness of LiDAR SLAM.
% \todo{high-level summary of evaluation-dtection-filtering}\bo{fixed}
In this section, we introduce a unified framework to evaluate LiDAR-based odometry and localization systems under common data corruptions, and we describe two defense strategies: a detection-and-filter pipeline and model fine-tuning on augmented data. 
% \triet{You should rename the subsections to something that describe the method/action, the current one sounds too results-focused. Something like "PCD Corruptions Modeling" or "Mitigation Methods"}
\begin{table}[]
    \centering
    \small
    \begin{tabular}{l|l}
        \toprule
        Category & Perturbation Type \\
        \midrule
        Weather (5) & rain, snow, rain + wet ground (rain\_wg), \\ 
        &snow + wet ground (snow\_wg), fog \\
        \midrule
        Noise (8) & background noise (bg\_noise), upsample, \\
         & uniform noise in CCS (uni\_noise), \\ 
         & gaussian noise in CCS (gau\_noise), \\ 
         & impulse noise in CCS (imp\_noise), \\ 
         & uniform noise in SCS (uni\_noise\_rad), \\ 
         & gaussian noise in SCS (gau\_noise\_rad), \\ 
         & impulse noise in SCS (imp\_noise\_rad), \\
        \midrule
        Density (5) & local density increase (local\_inc), \\ 
        & local density decrease (local\_dec), \\ 
        & beam deletion (beam\_del), \\
        & layer deletion (layer\_del), cutout \\
        \bottomrule
        
    \end{tabular}
    \caption{Our robustness framework \toolS consists of 18 real-world  LiDAR corruptions. CCS refers to Cartesian Coordinate System and SCS refers to Spherical Coordinate System. Their calculation formulas are listed in Table~\ref{tab:noise_def} and~\ref{tab:density_def}.}
    \label{tab:corruptions}
\end{table}

\subsection{\tool: A Robustness Framework for LiDAR Odometry and Localization}
\label{method:corruption_list}
Autonomous robots and vehicles are deployed in complex and diverse environments.
The performance of LiDAR may degrade under adverse conditions. 
To rigorously evaluate the robustness of odometry and localization systems, our framework employs 18 different types of PCD corruption (Table~\ref{tab:corruptions}) that simulate adverse weather, internal and external noises, 
% \todo{the following two are not described in this section}\bo{They are described in density perturbations}
equipment malfunctions, and occlusions. 
%from previous works~\cite{kilic2021lidar, HahnerICCV21, 10286105, HahnerCVPR22}. 
% We select a comprehensive set of corruptions reflecting commonly occurring phenomena that impact the LiDAR scans of the scenes.
% Next, we will describe these corruption in details.

\pa{Weather Perturbations}.
Adverse weather can negatively affect the quality of LiDAR scans. 
% The PCD recorded by LiDAR can change depending on the weather. 
% Droplets from rain, fog, and snow can alter the resulting scan, creating noises compared to identical scans under clear conditions. 
Randomly distributed airborne droplets from rain, snow, and fog scatter light, corrupting the PCD captured by LiDAR. 
Furthermore, precipitation results in wet ground that always co-occur with rain or snow, leading to surface reflections that differ significantly from dry ground.
Despite the availability of real-world datasets capturing various adverse weather conditions~\cite{ithaca, RobotCarDatasetIJRR}, we opt for simulation using clean datasets for three reasons.
% \todo{this bullets can be summarized and shortened, remove the itemize}\triet{done}
First, real datasets does not encompass the full spectrum of point-cloud corruptions, whereas simulation enables us to apply every corruptions on the same dataset, ensuring a comprehensive and fair comparison. Second, real recordings are sensor-specific, which can restrict compatibility with certain odometry or localization algorithms \cite{Zhang20243DLS}; simulation, by contrast, is sensor-agnostic and can be applied to any dataset. Finally, the severity of weather effects in real recordings is fixed and often undocumented, but simulation lets us precisely parameterize corruption intensity and systematically study its impact, leveraging realistic models of fog, rain, and snow \cite{10286105,kilic2021lidar,HahnerCVPR22,HahnerICCV21}.
% \begin{itemize}
    % \item \textbf{Realism.} Simulators model droplets in the air and simulate the propagation and scattering of lasers, which highly resembles adverse weather effects in the real world~\cite{10286105, kilic2021lidar, HahnerCVPR22, HahnerICCV21}.
    % \item \textbf{Controllability.} In real-world datasets, the severity of adverse weather is typically unknown and cannot be adjusted. Simulation, on the other hand, allows us to parameterize the intensity of these conditions. 
    % \item \textbf{Sensor limitations.} Real-world datasets are limited by their specific sensors, which may be incompatible with certain algorithms~\cite{Zhang20243DLS}. Simulation allows us to apply adverse weather on any dataset. 
    % \item \textbf{Completeness.} No dataset contains the full range of corruptions. Simulation allows us to apply all corruptions on a single dataset for a comprehensive and fair comparison. 
% \end{itemize}

\begin{figure}
    \centering
    \includegraphics[width=1\linewidth]{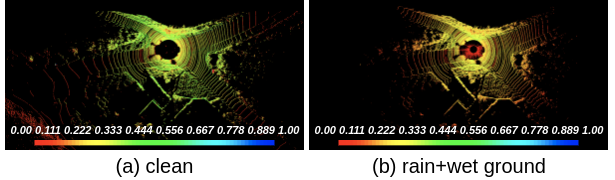}
    \caption{An example of rain and wet ground corruption. The field view and reflection intensities are significantly decayed compared to the clean data. The numbers above the color bar show the intensities of the points with corresponding colors.}
    \label{fig:rain_wg}
\end{figure}
Our framework simulates the following weather conditions considering both the direct impact of weather on LiDAR (e.g., droplets causing scattering), as well as impacts of wet grounds on LiDAR (e.g., reflections). 
\begin{itemize}
    \item \textbf{Rain or Snow}. We use LISA~\cite{kilic2021lidar, hegde2023source}, an off-the-shelf rain and snow simulator which use a physics model to simulate the decayed return of laser pulses during rain and snow.

    \item \textbf{Fog.} Following previous work~\cite{HahnerICCV21}, we model fog by altering the impulse response, capturing the effect of varying fog densities on the surrounding objects and environment.
    % While LISA already has a fog perturbation, it is more general, and works such as FS provide a better fog simulation model without needing unavailable data such as optical channel information. Hence, we use FS for fog simulation. 
    % FS simulates the effect of fog on real-world LiDAR point clouds that have been recorded in clear weather. 
    % FS models fog by modifying the impulse response changes to reflect the impact of different fog densities on the environment and the surrounding objects. 
    \item \textbf{Rain or Snow + Wet Ground (\textit{rain/snow\_wg})}. We combine LISA and Wet Ground Model (WGM)~\cite{HahnerCVPR22} to produce a more realistic simulation than rain/snow alone. WGM builds an optical model to simulate wet ground reflection based on water thickness. We show an example of this corruption in Figure~\ref{fig:rain_wg}.
%Note that We do not apply the Wet Ground Model standalone, instead, we apply it with rain and snow simulation. 
    % \item \textbf{Snow + Wet Ground (\textit{snow\_wg})}. Similar to rain + wet ground, we produce more realistic snow weather by combining LISA's snow simulation and the wet ground model on PCD.
\end{itemize} 
%we adopted three weather simulators for PCD, i.e., LiDAR Light Scattering Augmentation (LISA) for rain and snow, 
% LISA is a physics-based simulator that can add the effects of adverse weather conditions such as snow and rain on the PCD. LISA models droplets' effect on PCD and is 

% For wet ground simulation, we do not apply it standalone, instead, we combine it with rain and snow simulation. 

% \jinqiu{Add an example of perturbation with Rain+WetGround}\bo{It cannot be reflected by 2D visualization, it mainly changes the intensities, while visualization only cares if there is a point or not.}
% \triet{Bo pls revise the corruptions since you are more familiar}

\pa{Noise Perturbations}. 
Noises occur in LiDAR measurement due to internal (e.g., equipment vibration) or external (e.g., dust or strong light) factors~\cite{denoise_review, 6291650}. % Deng:17 
We design perturbations on point cloud to simulate the errors attributed to such factors. 
% We consider two kinds of noise modes in general. 
% The first one adds noises to the positions of existing points to simulate the errors happening in the measurement. 
\begin{table}[]
    \centering
    {\small
        \begin{tabular}{l|l}
        \toprule
            Noise Type &  Formula to Generate Noise\\\hline
           Gaussian (CCS)  & \mathsmall{$p_i +\Delta p * SEV,\; \Delta p \sim \mathcal{N}(0, 1)^3, p_i\in P$}\\\hline
           Uniform (CCS) & \mathsmall{$p_i +\Delta p * SEV,\; \Delta p \sim \mathcal{U}(-1, 1)^3, p_i \in P$} \\\hline
           Impulse (CCS) & \mathsmall{$p_i +\Delta p * C,\; \Delta p \sim \mathcal{U}\{-1, 1\}^3,$} \\ 
           & \mathsmall{$p_i \in RS(P, SEV)$} \\\hline
           Gaussian (SCS)  & \mathsmall{$p_i +\Delta r * SEV,\; \Delta r \sim \mathcal{N}(0, 1), p_i\in P$}\\\hline
           Uniform (SCS) & \mathsmall{$p_i +\Delta r * SEV,\; \Delta r \sim \mathcal{U}(-1, 1), p_i \in P$} \\\hline
           Impulse (SCS) & \mathsmall{$p_i +\Delta r * C,\; \Delta r \sim \mathcal{U}\{-1, 1\},$} \\ 
           & \mathsmall{$p_i \in RS(P, SEV)$} \\\hline
           Background   & \mathsmall{$P_{new} = P \cup \{p_i'|j=1, 2, ..., SEV\}$}\\
           (CCS) & \mathsmall{$p_i'=\{x_i', y_i', z_i'\}, 
    \left\{ 
    \begin{array}{l}
    x_i' \sim \mathcal{U}(x_{\min}, x_{\max}) \\
    y_i' \sim \mathcal{U}(y_{\min}, y_{\max}) \\
    z_i' \sim \mathcal{U}(z_{\min}, z_{\max}) \\
    \end{array} 
    \right.$}\\\hline
           Upsample & \mathsmall{$P_{new} = P \cup \{p_i + \Delta p| \Delta p \sim \mathcal{U}(-0.1, 0.1),$}\\
           (CCS) & \mathsmall{$ p_i\in RS(P, SEV)\}$} \\
        \bottomrule
        \end{tabular}
        \caption{Noise-induced corruptions. \textit{SEV} (severity) is the significance of corruption, whose definition varies based on the type of corruption. \textit{C} is a constant value, whose concrete value varies in different scenarios. \textit{RS} is a random subset function, which samples a subset from the input based on the given $SEV$. $\Delta p$ is a point in CCS, so $+\Delta p$ means we change the three coordinates in CCS. $\Delta r$ refers to the range value, $+\Delta r$ means we only change the range value in SCS.} 
        \label{tab:noise_def}
    }
\end{table}

We show the details of noise perturbations in Table~\ref{tab:noise_def}. One frame of point cloud is annotated as $\mathcal{P} = \{p_i|i=1, 2, ... N \}$, where $N$ is the number of points in the point cloud. We apply three distributions of noises (Gaussian, Uniform, and Impulse) to two coordinate systems (Cartesian and Spherical, CCS and SCS for short) to simulate different noise sources. 
%There exist two coordinate systems (CS) that are commonly used by SLAM techniques to represent the PCD points, namely Cartesian (CCS) and Spherical coordinate systems (SCS). 
In CCS, one point is represented by $p=\{x, y, z\}$, i.e., Euclidean distance to each planes, while in SCS, one point is represented by $p=\{r, \phi, \theta\}$, where $r$ is the range, i.e., Euclidean distance from the origin to the point, $\phi$ is the azimuthal angle and $\theta$ is the polar angle. 
Noises in CCS represent inaccuracies caused by the vibration or rotating of the sensor~\cite{dong2023benchmarking} while SCS can better represent inaccurate measurement of time of flight (ToF)~\cite{10286105}.
In addition to re-positioning existing LiDAR points, we synthesize new points to simulate environment noises, such as dust in the air or strong external light, through \textit{background noise} and \textit{upsample} applied to CCS. Background noise adds points randomly in the entire space, while upsample adds points close to randomly selected points. 

\pa{Density Perturbations}.
\begin{table}[]
    \centering
    \begin{tabular}{l|l}
    \toprule
        Corruption &  Formula to Generate Noise\\\hline
        Local Inc &  \mathsmall{$P_{new} = P \cup P_{add}, $}\\
        & \mathsmall{$P_{add}=\bigcup_{i=0}^{SEV}IP(NN_{100}(p_i))$} \\\hline
        Local Dec &  \mathsmall{$P_{new} = P \setminus P_{rm},$} \\
        & \mathsmall{$P_{rm} = \bigcup_{i=0}^{SEV}RS(NN_{100}(p_i), C)$} \\\hline
        Cutout &  \mathsmall{$P_{new} = P \setminus \bigcup_{i=0}^{SEV}NN_{20}(p_i)$} \\\hline
        Beam Del &  \mathsmall{$P_{new} = RS(P, SEV)$} \\\hline
        Layer Del &  \mathsmall{$P_{new} = P \setminus \bigcup_{i\in S}Layer_i,$} \\
        & \mathsmall{$S=RS(\{1,2,...64\}, SEV), P = \bigcup_{i=1}^{64} Layer_i$} \\
    \bottomrule
    \end{tabular}
    \caption{Density-related corruptions. \textit{SEV} (severity) refers to the significance of corruption, whose definition varies based on the type of corruption. \textit{C} refers to a constant value, whose concrete value varies in different places. \textit{IP} refers to the 3D interpolation function. $NN_k$ refers to the k-nearest neighbor function. \textit{RS} refers to a random subset function, which samples a subset from the input based on the given number.}
    \label{tab:density_def}
\end{table}
The density of points can be changed due to internal (e.g., malfunction of the equipment) or external (e.g., occlusion between objects and reflection on object surface) factors~\cite{10588461, xu2022behind}. 
We simulate five density-related corruptions, local density increase, local density decrease, cutout, beam deletion, and layer deletion. 
The definitions of these corruptions are listed in Table~\ref{tab:density_def}.

\subsection{Improving the Robustness of LiDAR Odometry and Localization}\label{sec:improvement}
We propose and experiment with two types of approaches to improve the robustness of such LiDAR systems. 
\subsubsection{For All LiDAR Odometry and Localization.}
\label{method:det_and_filter}
We propose a novel detection-and-filter pipeline to improve the robustness of LiDAR odometry and localization systems. This pipeline is algorithm-agnostic and can be widely applied to algorithms of LiDAR pose estimations. 
%Classical (non-learning) systems lack the flexibility to adapt to new data distributions without manual algorithmic updates. To provide a general-purpose defense, we propose a two-stage detection-and-filter 
Our novel pipeline consists of two stages: (1) corruption detection via a DNN-based classifier, and (2) targeted corruption filtering based on the corruption type. 

\textit{Stage 1 -- Corruption Detection.}
As odometry and localization are real-time applications, we focus on techniques that are inexpensive in terms of time and memory cost. Hence, we first convert each 3D LiDAR scan into a 2D range image via spherical projection similar to \cite{9561063}, creating a $64 \times 1024$ grid, where each pixel stores the depth and intensity of the point in that cell closest to the agent.
This representation preserves the spatial structure of the scene while enabling efficient convolutional processing.

\begin{figure}
    \centering
    \includegraphics[width=\linewidth]{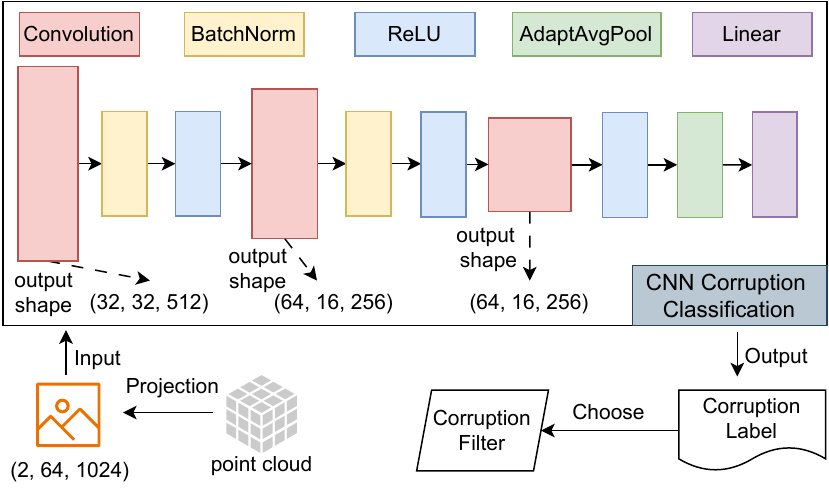}
    \caption{Architecture of CNN for corruption classification.
    % \jinqiu{vertical...very difficult to read; fix it to three piles of stacked horizontal boxes and connect them}
    }
    \label{fig:noise_classifier_cnn}
    
\end{figure}

Using these range images, we train a lightweight CNN (Figure \ref{fig:noise_classifier_cnn}) to classify data corruptions (e.g., rain, noises) in LiDAR data on our augmented dataset by minimizing categorical cross-entropy loss. This loss guides the model to improve classification accuracy across different corruption types through successive training epochs. 

The CNN consists of three convolutional layers with increasing feature sizes (32, 64, and 128), each followed by batch normalization and ReLU activation. A final adaptive average pooling layer reduces the output to a 128-dimensional vector to produce class scores, where the class with highest probability is selected as the predicted class for the filtering stage.
% The network outputs a probability distribution over the corruption classes:
% \[
% \hat{\mathbf{y}} = \mathrm{softmax}\bigl(f_{\theta}(\mathbf{I})\bigr), \quad \mathbf{I} \in \mathbb{R}^{H\times W},
% \]
% where $f_{\theta}$ denotes the CNN parameters. 
% corruption type $= \arg\max_c \hat{y}_c$ and proceed to the filtering stage.
This lightweight architecture is designed for real-time or embedded deployment, balancing classification accuracy with minimal computational overhead. 

\textit{Stage 2 -- Corruption Removal.}
After identifying the corruption type, we apply the corresponding filter to mitigate its effects and restore point-cloud fidelity. 
While our framework supports any paired detector–filter module (e.g., precipitation or density filters), in this work we concentrate on noise-induced corruptions because other point-cloud-based systems are sensitive to noises \cite{10286105} and denoising methods are mature and effective in removing noises.
Thus, once the classifier flags a scan as noise-corrupted, we pass it to a bilateral filter~\cite{ipol.2017.179} for denoising.
This modular design allows additional filters to be plugged in for other corruption categories.

\subsubsection{For Learning-Based LiDAR Odometry and Localization.}
Learning-based approaches inherently adapt to the characteristics of their training data. 
Hence, to defend these systems against the data corruptions, we augment the training set with perturbations shown to degrade performance by our evaluation framework and then fine-tune the pretrained models on this augmented dataset. 
We fine-tune the model using the original training recipe from the released checkpoint until it converges on the augmented dataset. 
This strategy encourages the neural networks to learn invariant features directly from the data distribution, and has been shown to enhance resilience in LiDAR-based systems~\cite{kilic2021lidar, HahnerCVPR22, HahnerICCV21}.
% \todo{include fine-tuning details. move from section 5.3 to  here and remove any overlap texts}

\section{Experiment Setup}
\subsection{Subject LiDAR Pose Estimation Systems}
\begin{table}[ht]
  \centering
  {\small
    %     \setlength{\tabcolsep}{1.5mm}
    %   \begin{tabular}{llllr}
    %     \toprule
    %     Category & Method & Year & Method & Error \\
    %     \midrule
    %     Odometry & MULLS & 2021 & Feature-based      & 0.56\% \\
    %     & KISS-ICP  & 2023 & Direct      & 0.51\% \\
    %     & DeLORA & 2021  & Learning-based & 7.78\% \\
    %     & NeRF-LOAM & 2023 & Learning-based & 1.70\%  \\
    %     \midrule
    %     Localization & LocNDF & 2023 & Learning-based  & 0.059m\\
    %   \bottomrule
    % \end{tabular}

      \begin{tabular}{llll}
        \toprule
        Category & Name & Year & Method \\
        \midrule
        Odometry & MULLS & 2021 & Feature-based \\
        & KISS-ICP  & 2023 & Direct \\
        & DeLORA & 2021  & Learning-based \\
        & NeRF-LOAM & 2023 & Learning-based  \\
        \midrule
        Localization & LocNDF & 2023 & Learning-based \\
      \bottomrule
    \end{tabular}
  }
    % \caption{Subject LiDAR odometry and localization techniques. We reported the errors based on our experiments, which are very similar to the numbers in their original papers.}
    \caption{Subject LiDAR odometry and localization techniques.}
  \label{tab:slams}
\end{table}
We evaluate four odometry and one localization systems, as shown in Table~\ref{tab:slams}.
The subject systems comprise two representative classical methods, MULLS and kiss-icp, and three learning-based approaches, DeLORA, NeRF-LOAM, and LocNDF.
MULLS~\cite{9561364} is a feature-based LiDAR odometry that uses ground filtering and principal components analysis to extract features from the point cloud.
It builds local sub-maps using the previous point cloud and match the current frame with local sub-maps by the proposed multi-metric linear least square iterative closest point algorithm to compute the pose.
KISS-ICP~\cite{Vizzo2022KISSICPID} is a direct matching LiDAR odometry that combines point-to-point ICP with adaptive thresholding, a robust kernel, universal motion compensation, and point cloud subsampling.
% ~\triet{Bo pls correct this part when you can since you are more familiar}
DeLORA~\cite{9561063} is a learning-based, unsupervised LiDAR odometry.
It projects the 3D point cloud to 2D image representations and employs ResNet to extract features from the 2D image.  
Then it predicts the translation and rotation by feeding the features into a neural network. 
NeRF-LOAM~\cite{Deng_2023_ICCV} employs a neural network to predict the signed distance field (SDF). 
It first builds the previous frames into an octree. 
% Then it samples some points from the laser rays which intersect with the voxels in the previously built octree. 
Then it samples intersected points between laser rays and voxels in the previously built octree.  
Given sampled points, the neural network predicts the SDF. 
Since sampled points are transformed to the same coordinate system as the previously built octree by the pose to be estimated, NeRF-LOAM optimizes the pose, neural network, and voxel embedding jointly by minimizing the loss between predicted SDF and real SDF. 
Different from NeRF-LOAM, LocNDF trains a neural network offline to predict the SDF. 
Given the online LiDAR scan, it tries to find a pose $P$ so that the neural network can produce minimal SDF when sampled points are transformed by $P$.  

% In addition to traditional SLAMs, we also include learning-based SLAMs in our evaluation. 
% While learning-based SLAM methods have not reached the accuracy and reliability of traditional methods, their ability for unsupervised learning without expertly designed algorithms facilitates their gaining popularity.
% Delora~\cite{9561063} is a projection-based, self-supervised, learning-based~\todo{describe}.
% LocNDF~\cite{wiesmann2023ral} trains a neural network to learn a discretization-free distance field of a given scene where new scans can be directly registered for precise localization. 
% NeRF-LOAM~\cite{Deng_2023_ICCV} is a neural radiance field (NeRF)-based SLAM that facilitates online learning and requires not pre-trained models~\todo{describe}.

% The chosen localization methods reflect the three sub-categories of LiDAR SLAM, as well as methods using both handcrafted and learning-based features.

\subsection{Datasets}
\label{exp:datasets}
%\begin{table}[ht]
%  \caption{Datasets}
%  \label{tab:datasets}
%  \centering
%  \scalebox{1}{
%  \begin{tabular}{llr}
%    \toprule
%    Dataset & Year & Testing Frames\\
%    \midrule
%    KITTI 09~\cite{Geiger2012CVPR} 
%            & 2012 & 1591\\
%    KITTI 10~\cite{Geiger2012CVPR} 
%            & 2012 & 1201\\
%    ColumbiaPark~\cite{L3NET_2019_CVPR}     
%            & 2019 & 700\\
%
%  \bottomrule
%\end{tabular}
%}
%\end{table}
% We use two different datasets for this study. 
% \bo{Add motivation for dataset selection}
We use KITTI Visual Odometry/SLAM Evaluation 2012 (KITTI)~\cite{Geiger2012CVPR}
to evaluate the four odometry systems, DeLORA, MULLS, KISS-ICP, and NeRF-LOAM. 
The reason is that KITTI is one of the most popular dataset for odometry and all of the selected odometry systems have been evaluated on KITTI.
KITTI 00 to 10 sequences are released with ground truths. 
We conducted our experiment on sequences 09 and 10 because (1) DeLORA was trained on sequences 00-08 hence only 09 and 10 remain. 
% which contain ground truth to measure the errors resulting from perturbations; 
(2) sequences 09 and 10 are of moderate sizes which facilitate the large number of perturbations in a reasonable amount of time and computation resources.
% The four methods Delora, MULLS, KISS-ICP, and NeRF-LOAM are evaluated on this dataset.

We use Apollo South-Bay~\cite{L3NET_2019_CVPR} to evaluate LocNDF.
Apollo South-Bay contains completed ADS recordings covering various driving scenarios.
LocNDF released the pretrained model and pre-built map on this dataset. 
In this work, we used a subset of ColumbiaPark-3, consisting of 700 testing frames, which is 
% This includes 800 frames for training, 800 frames for validation, and 700 frames for testing. 
% We use this subset to make our evaluation 
consistent with LocNDF's pre-trained model~\cite{wiesmann2023ral}. 
% The extensive collected data allows for separation between training, validation, and testing and the construction of a map allowing LocNDF to be evaluated on this dataset.

\subsection{Evaluation Metrics}
% \todo{I don't find rot metric is used in the paper, if not used, then it can be removed; what is a `pose'? this should be included, is this a xyz coordiantor or what? what is the unit?}\bo{Fixed}
For the evaluation of the precision of pose estimation, we adopted the translational relative pose error (\rpetrans{})~\cite{Geiger2012CVPR}. 
\rpetrans{} is widely used in previous work~\cite{Zhang20243DLS} to measure and compare the performance of pose estimation. \rpetrans{} is calculated as:
% \begin{equation}
% RPE_{rot}(\mathscr{F}) = \frac{1}{|\mathscr{F}|} \sum_{(i, j) \in \mathscr{F}} \angle [(\hat{p}_j \ominus \hat{p}_i) \ominus (p_j \ominus p_i)]
% \end{equation}
\begin{equation}
RPE_{trans}(\mathscr{F}) = \frac{1}{|\mathscr{F}|} \sum_{(i, j) \in \mathscr{F}} ||(\hat{p}_j \ominus \hat{p}_i) \ominus (p_j \ominus p_i)||_2
\end{equation}
% where 
% $RPE_{trans}$ measures the relative translation error and $RPE_{rot}$ measures the relative rotation error between the estimated pose and the ground truth, 
% $\mathscr{F}$ is a set of frames $(i, j)$, $\hat{p}$ is the estimated pose and $p$ is the ground truth pose, $\ominus$ is the inverse compositional operator~\cite{Kmmerle2009OnMT}.
% and $\angle[\cdot]$ is the rotation angle.
where $\mathscr{F}$ is a set of frames $(i, j)$, $\hat{p}$ is the estimated pose and $p$ is the ground truth pose, $\ominus$ is the inverse compositional operator~\cite{Kmmerle2009OnMT}. A pose $p$ consists of a rotation matrix $R$ and a translation vector $t$.  

% \subsection{Calculation of the error due to perturbations}
% For each SLAM method, before we can evaluate the effect of the perturbations, we establish a baseline by running it on the original dataset (Table~\ref{tab:datasets}) and record the RPE.

% Hence, for each perturbation, we would calculate the impact via the change in RPE relative to the baseline. \bo{We did not actually report this, we reported the errors in each situation directly}. 

\section{Experiment Results}
% \todo{Missing the results of locNDF?}\bo{LocNDF is not impacted by any corruption, there is no performance change under any corruption. Added the description in the beginnign of the results.}
%\subsection*{\RQOne}
\subsection{\tool: Robustness Results of LiDAR Odometry and Localization}
\label{exp:rq1}
%\pa{Motivation}. While the effects of PCD perturbations on 3D obstacle detection are widely studied, the effects of the same perturbations on vehicle localization are unclear. Yet, in the context of an ADS, they utilize identical data. Hence, we are interested in learning the impact of such perturbations on localization and determining whether the perturbations that have shown significant impacts on obstacle detection can also impact localization.
% \begin{figure}
%     \centering
%     \includegraphics[width=1\linewidth]{images/rpe_trans_v2.png}
%     \caption{Performance of each system under selected corruptions on KITTI 10. Eight representative corruptions are shown due to space constraints.}
%     \label{fig:rq1_results}
% \end{figure}
\pa{Method}. We applied our robustness framework to evaluate the subject odometry and localization systems. 
% The robustness framework transforms
We first perturb the KITTI and Apollo-SouthBay datasets
% based on the perturbations 
with the corruptions in Table~\ref{tab:corruptions}. %Then, we applied the subject SLAMs(Table~\ref{tab:slams}) are used to evaluate the perturbed data.
% We report the results in RPE, a metric commonly used in SLAM. 
Note that NeRF-LOAM is computationally expensive (i.e., requiring at least 24 GB GPU memory and several days to finish a single sequence), so we took a statistically-significant sample of frames from KITTI for NeRF-LOAM as a proof of concept.
\begin{figure*}[!htbp]
    \centering
    \includegraphics[width=0.9\linewidth]{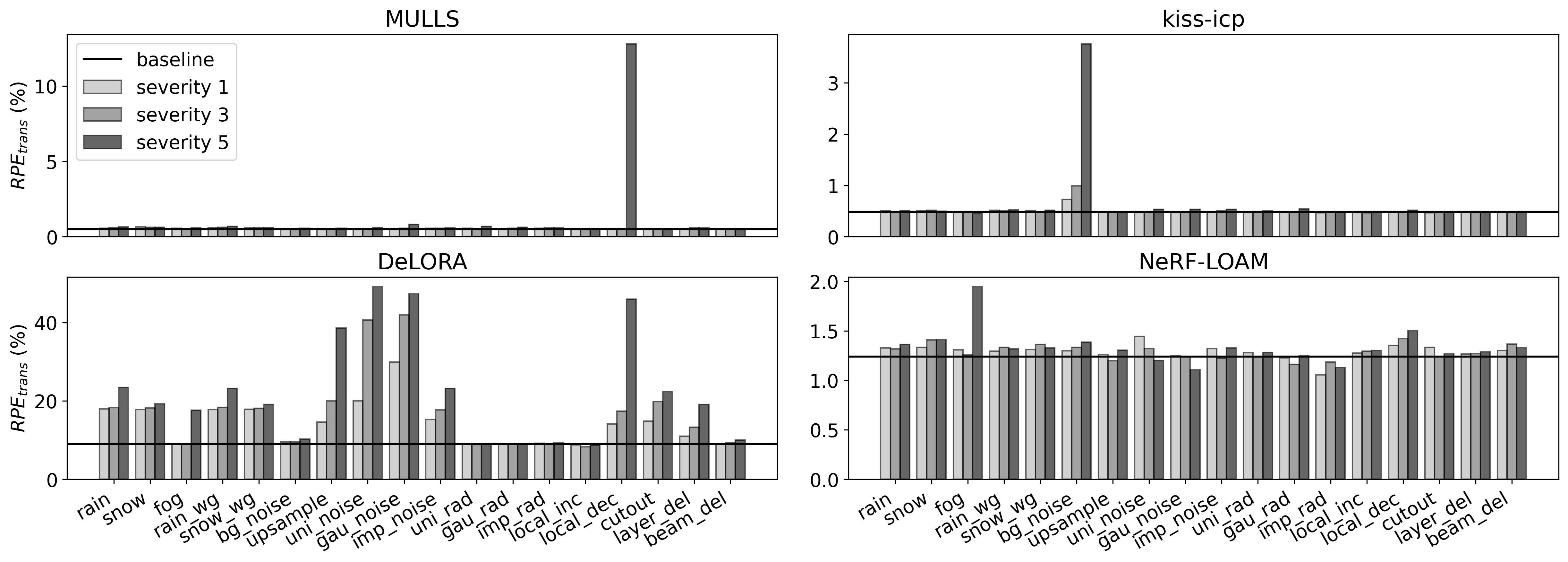}
    \caption{Performance of each system on KITTI 09.}
    \label{fig:seq09_full}
\end{figure*}
\begin{figure*}[!htbp]
    \centering
    \includegraphics[width=0.9\linewidth]{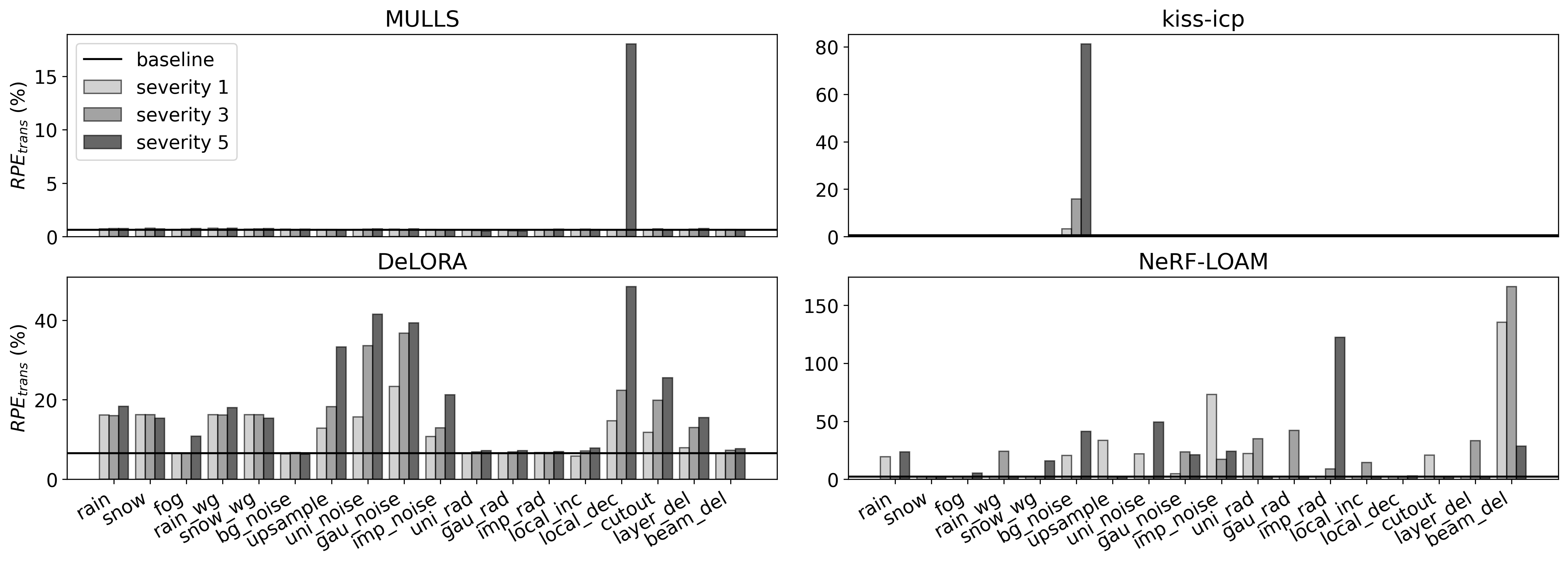}
    \caption{Performance of each system on KITTI 10.}
    \label{fig:seq10_full}
\end{figure*}

\pa{Results}. 
Figure~\ref{fig:seq09_full} and~\ref{fig:seq10_full} present \rpetrans{} on KITTI 09 and 10 for all corruptions of severities 1, 3, and 5. 
We omit the results of \textbf{\textit{LocNDF}} as we observe no performance degradation under any corruption in our experiment. 
We share the complete results in the released artifacts. 
% We show the results of \rpetrans{} in Figures~\ref{fig:rq1_results}. 
% Due to the space limitation, we only present the results on KITTI 10 for a subset of corruptions at severity 1, 3 and 5, which cover all the vulnerability exposing corruptions. 
% We do not present the \textit{RPE}$_{rot}$ in the paper as it is correlated with \rpetrans. 
% We do not present the results of \textbf{\textit{LocNDF}} because it is robust and there is no performance change under any data corruption.
% We share the full detailed results of \rperot{} in our public artifact.  

Non-learning systems show overall robustness to most perturbations with each system particularly sensitive to certain corruption types, leading to significant performance drops. \textbf{\textit{MULLS}} is sensitive to \textit{local density decrease (local dec)} at severity 5 where \rpetrans{} increases from 0.3\% to over 18\% and \textbf{\textit{kiss-icp}} is sensitive to \textit{background noise} where \rpetrans{} rises sharply from less than 0.6\% on the clean data to over 80\%. Our investigation shows that local density reduction disrupts key point clusters which \textbf{\textit{MULLS}} relies on, making its feature extraction unreliable. On the other hand, \textit{background noise} introduces uniformly distributed noisy points which disrupt \textbf{\textit{kiss-icp}}'s direct point-on-point registration.

% For example, while \rpetrans{} for \textbf{\textit{MULLS}} increases less than 0.3\% under most corruptions \textbf{\textit{MULLS}}, it spikes to over 18\% when \textit{local density decrease} (severity 5) is applied.
% However, with \textit{local density decrease} (severity 5), the error of \textbf{\textit{MULLS}} spikes to over 
% 12\% on KITTI 09 and 18\% on KITTI 10. 
% 18\%.
% This is because \textbf{\textit{MULLS}} relies on geometric features (e.g., planes and edges) for registration, and local density reduction disrupts key point clusters, making feature extraction unreliable.
% Similarly, \textbf{\textit{kiss-icp}} is particularly vulnerable to \textit{background noise}. 
% Its \rpetrans{} rises sharply from less than 0.6\% on the clean data to over 80\% with \textit{background noise} at severity 5,
% on KITTI 10, 
% while the error only increases by less than 0.2\% in most cases. 
% Even at severity 1, \textit{background noise} causes a notable increase, from under $0.6\%$ to more than $3\%$ \rpetrans
% ~on KITTI 10. 

% This is because \textbf{\textit{kiss-icp}}'s direct point-on-point registration and \textit{background noise} introduces uniformly distributed noisy points which are unable to be matched across the scene. 

In contrast, learning-based odometry systems, i.e., \textbf{\textit{DeLORA}} and \textbf{\textit{NeRF-LOAM}} are more sensitive to most corruptions. 
% For instance, the \rpetrans{} of DeLORA increases significantly, from below 10\% to as high as 50\%, under many corruptions. 
Notable exceptions include background noise, spherical-coordinate noises (uniform, Gaussian and impulse noises (SCS)), local density increase, and beam deletion, where performance remains relatively stable due to \textit{DeLORA}'s 2D range image representation.
% and CNN-based feature extraction. 
% In this representation, 
% where \textit{background noise} appears as mild pixel-level noise, which CNNs are inherently robust to. 
% During projection, multiple 3D points may collapse into the same pixel location, with only the closest point retained. 
% As a result, \textit{spherical-coordinate noise} primarily alters depth values without changing pixel locations; \textit{beam deletion} randomly deletes points, however, redundancy ensures minimal impact; and \textit{local density increase} adds additional points in a cluster that often map to existing pixels, preserving the image structure. 
% By contrast, other corruptions introduce geometric distortion or loss of key features in the image, leading to the severe performance drops observed. 
During projection, multiple 3D points may map to the same pixel, with only the closest point retained, which significantly limit the effect of these perturbations. Thus, \textit{background noise} appears as mild pixel-level noise, \textit{spherical-coordinate noise} alters depth without affecting pixel locations, \textit{beam deletion} has limited impact due to point cloud redundancy, and \textit{local density increase} adds points that often overlap with existing pixels, preserving image structure. In contrast, other corruptions introduce geometric distortions or remove key features, leading to more severe performance drops.
Finally, we find that \textbf{\textit{NeRF-LOAM}} shows substantial instability, particularly on KITTI 10. 
In some cases, increasing corruption severity produces lower errors which is counterintuitive. 
To investigate this, we repeated the experiments under each corruption two additional times (three runs in total) with identical inputs. 
The resulting trajectories and final errors vary substantially across runs. 
We attribute this variability to random network initialization and the limited number of online optimization iterations. 

\rqboxc{Our systematic evaluation reveals key vulnerabilities in various LiDAR odometry systems.}

\subsection{Detection-and-Filter Pipeline}\label{sec:rq2.1}
\begin{figure*}
    \centering
    \includegraphics[width=1\linewidth]{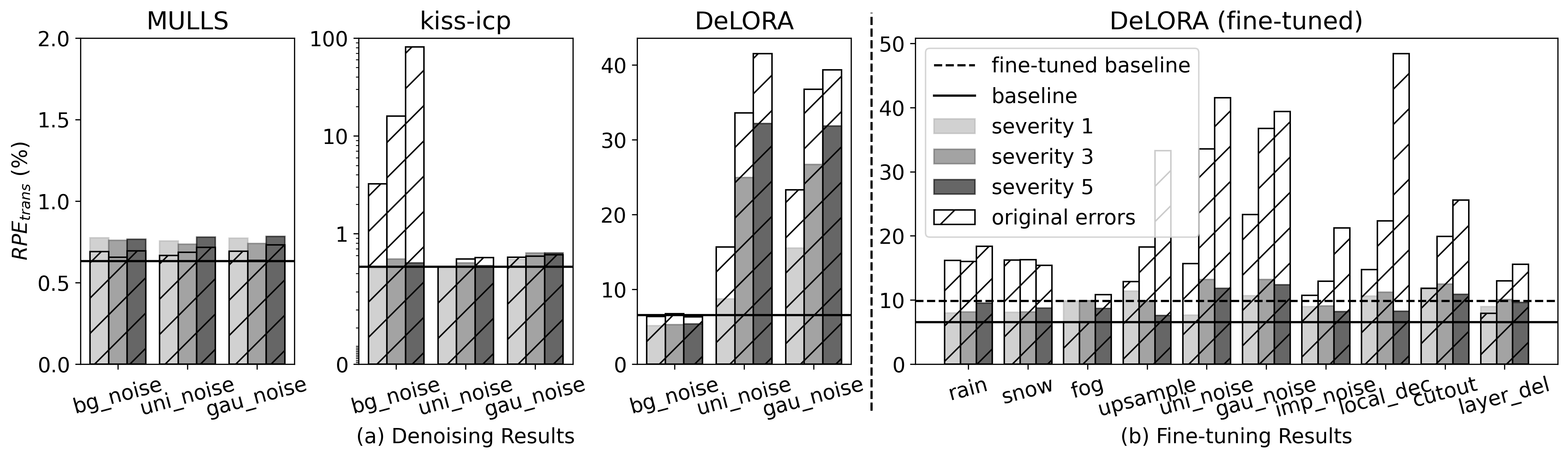}
    \caption{(a) $RPE_{trans}$ with our detection-and-filter pipeline. (b) $RPE_{trans}$ of the fine-tuned DeLORA on the augmented dataset. Colored bars (light gray, medium gray, dark gray) report the RPE at severity levels 1, 3, and 5 after applying our pipeline or fine-tuning; the hatched white bars with ``/'' show the corresponding errors without our pipeline or fine-tuning.}
    \label{fig:denoise}
\end{figure*}
\pa{Method}. We apply the detection-and-filter pipeline on all systems except 
\textbf{\textit{NeRF-LOAM}} because it could not produce consistent results between different trials on the exact same data and 
\textbf{\textit{LocNDF}} because it is robust against all proposed corruptions. 

% \textit{A. Corruption Detection.} 
We implement the classifier proposed in Section~\ref{method:det_and_filter} using PyTorch 1.13.1+cu117 and train it to recognize eight representative corruptions: rain, snow, fog, uniform noise (CCS), Gaussian noise (CCS), background noise, local density increase, and local density decrease. 
Due to their similar scattering effects, we merge rain and snow into one class; likewise, uniform and Gaussian noise (CCS) are grouped together for their comparable distortion patterns.
We train the model on KITTI 09 (80\% train, 20\% validation) for 10 epochs using a batch size of 8, a learning rate of 0.001 with the Adam optimizer. 
We apply it on KITTI 10 and we use the bilateral filter to restore the noise-corrupted PCD. 
All experiments are run on Ubuntu 22.04 LTS with an AMD Ryzen 32-core CPU, 256 GB RAM, and an NVIDIA GeForce RTX 3090 GPU. 

% \textit{B. Corruption Filter.}
% For each corruption type, an appropriate filter restores point-cloud fidelity. 
% In this study we focus on noise-induced corruptions for three reasons: (i) our classifier achieves very high accuracy on identifying noisy point clouds; (ii) noises pose a significant security threat to MULLS, kiss-icp, and DeLORA; and (iii) the denoising literature offers well-established methods. We adopt a bilateral filter to remove noises while preserving structural detail.

\pa{Results}.
Our classification model achieves an average accuracy of 95.54\%, as shown in Table~\ref{tab:classification_results}. 
We notice that clean and fog have a comparably lower accuracy because they are often misclassified as local density increase. 
The reason is that (i) fog has sparse reflections which increase the points density, and (ii) occasional high-reflecting surfaces in clean data also increases points density. 
\begin{table}[ht]
  \centering
  {\small
        \setlength{\tabcolsep}{1.1mm}
      \begin{tabular}{l|r|rr|rr|rr}
        \toprule
         &       & \multicolumn{2}{c|}{Weather}           & \multicolumn{2}{c|}{Noise}            & \multicolumn{2}{c}{Density} \\
            & \multicolumn{1}{c|}{Clean} & \multicolumn{1}{c}{Fog}     & \multicolumn{1}{c|}{R+S} & \multicolumn{1}{c}{U+G} & \multicolumn{1}{c|}{Back}  & \multicolumn{1}{c}{Inc} & \multicolumn{1}{c}{Dec}\\

        \midrule
        % \multirow{4}{*}{\rotatebox[origin=c]{90}{}}
        % KITTI 09  & 99.94 & 99.42 & 100   & 100 & 100 & 100 & 98.91\\
        Acc  & 90.02\% & 80.83\% & 98.32\% & 100\% & 100\% & 100\% & 97.84\%\\
      \bottomrule
    \end{tabular}
  }
    \caption{Classification results for seven class of noises, i.e., Clean, Fog, Rain+Snow (R+S), Uniform+Gaussian Noise (U+G), Background Noise (Back), Local Density Increase (Inc), and Local Density Decrease (Dec)}
  \label{tab:classification_results}
\end{table}

% \begin{figure}
%     \centering
%     \includegraphics[width=1\linewidth]{images/rpe_denoise.png}
%     \caption{Comparison of $RPE_{trans}$ with and without detection-and-filter pipeline. Colored bars (light gray, medium gray, dark gray) report the RPE at severity levels 1, 3, and 5 after applying our pipeline; the hatched white bars with ``/'' show the corresponding errors without filtering.}
%     \label{fig:denoise}
% \end{figure}

We present the \rpetrans{} of the three systems incorporated with our pipeline on KITTI 10 in Figure~\ref{fig:denoise} (a).
% KITTI 09 has a similar result and is available in the appendix. 
\textbf{\textit{MULLS}} is shown to be robust against most noise in the previous section. 
Applying our detection-and-filter pipeline to the noisy data causes negligible performance degradation (less than 0.1\%). 
\textbf{\textit{kiss-icp}} significantly benefits from the detection-and-filter pipeline, with post-denoising errors returning to the clean-data baseline. 
\textbf{\textit{DeLORA}} also has a significant error reduction after filtering noises in data, however, the errors are still much higher than the baseline. 

\rqboxc{Our classifier could classify noise types accurately and our detection-and-filter pipeline could significantly reduce the error on noise-corrupted point cloud data. }

\subsection{Fine-Tuning}
\pa{Method}. We fine-tune the learning-based system, i.e., \textbf{\textit{DeLORA}}, on the dataset augmented with the corruptions that affect it most significantly, namely, rain, snow, fog, upsample, uniform noise (CCS), Gaussian noise (CCS), impulse noise (CCS), local density decrease, cutout, and layer deletion. 
We exclude \textbf{\textit{NeRF-LOAM}} and \textbf{\textit{LocNDF}} for the same reasons mentioned in Section~\ref{sec:rq2.1}. 
% We augment the training set (i.e., KITTI 00 to 08 sequences) of \textbf{\textit{DeLORA}} by applying the proposed point cloud corruptions and fine-tune it on the augmented data.
% More specifically, we fine-tune \textbf{\textit{DeLORA}} on the corruptions that affect it most significantly, namely, rain, snow, fog, upsample, uniform noise, Gaussian noise, impulse noise, local density decrease, cutout, and layer deletion. 
% We only augmented the training data with these corruptions at severity 5 and fine-tuned \textbf{\textit{DeLORA}} for 6 epochs with the default training hyperparameters.  

\pa{Results}.
% \begin{figure}
%     \centering
%     \includegraphics[width=1\linewidth]{images/rpe_retrain.png}
%     \caption{Comparison of $RPE_{trans}$ before and after retraining of DeLORA.}
%     \label{fig:retrain}
% \end{figure}
We present the performance of the fine-tuned \textbf{\textit{DeLORA}} on corrupted KITTI 10 scans (KITTI 09 results are available in the replication package) in Figure~\ref{fig:denoise} (b). 
We find that the errors are reduced by up to 35.79\% for all types of corruption after fine-tuning which brings the performance close to the clean data. 
On the baseline, the fine-tuned \textbf{\textit{DeLORA}} achieves a 2.7\% decrease in \rpetrans{} on KITTI 09 but a 3.3\% increase on KITTI 10 compared to the original model.  
It is important to note that our fine-tuning was comparably simple: we only augmented KITTI 00 to 08 with the selected corruptions at severity 5 and used the original learning rate. 
We believe that further improvements could be achieved by a more refined fine-tuning process, such as augmenting the dataset with the corruption at different severities, and adjusting the learning rate and the ratio between clean and corrupted data.  

\rqboxc{Fine-tuning with the corruption data is an effective way to improve the robustness of the learning-based LiDAR odometry system (DeLORA).}

% \input intro
% \input related
% \input approach
% \input setup
% \input eval
% \input threats
% \input conclusion

% \section{Discussion and Future Work}
% \bo{todo: update}
% In future work, we plan to extend our method to a two-stage classification process, where corruptions that are frequently classified as each other as grouped together in the first step (multi-class prediction) and then are separated using a binary classifier in the second step\bo{What are the benefits. If two classes cannot be distinguished in multi-class situations, they cannot be distinguished in binary situation either. The core is that they are too similar}. This enhance accuracy in the prediction and allow us to use specific noise alleviation strategy for each class. 
% Furthermore, we will extend our pipeline to repair the PCD via an encoder for corrupted scene reconstruction. This will enhance the method's application and allow generalization to all black-box systems.

% \bo{In this work, we only evaluated our detection-and-filter defense strategy on noise-induced corruptions. We will extend our work to other types of corruptions by incorporating other filters such as rain and snow removal.}
% \bo{Or we simply remove this section.}

\section{Conclusion}
% This work highlights the need the evaluate the robustness of LiDAR SLAM methods against commonly occurring LiDAR data corruption, especially for newly developed learning-based SLAMs. 
% Our results show that each LiDAR-based SLAM is vulnerable to different corruptions and this should be considered when selecting the SLAM method.
In this work, we propose \textbf{\textit{\tool}},  a systematic framework for evaluating the robustness of LiDAR-based odometry and localization systems against 18 types of synthetic real-world point cloud corruptions.
% Our findings indicate that all SOTA LiDAR-based odometry systems are particularly vulnerable to certain corruption types. 
Our evaluation reveals critical vulnerabilities in different types of odometry systems, with performance degradation of up to 80\%. 
We further demonstrate that our simple detection-and-filter pipeline can effectively mitigate noise-induced corruptions, restoring performance to near-clean levels. 
Finally, we show that corruption-based data augmentation substantially enhances the robustness of learning-based systems against point cloud corruptions. 
Notably, it also improved clean-data performance on KITTI 09, though a slight decrease was observed on KITTI 10.

% \section*{Acknowledgments}
% This should be a simple paragraph before the References to thank those individuals and institutions who have supported your work on this article.

% {\appendix[Proof of the Zonklar Equations]
% Use $\backslash${\tt{appendix}} if you have a single appendix:
% Do not use $\backslash${\tt{section}} anymore after $\backslash${\tt{appendix}}, only $\backslash${\tt{section*}}.
% If you have multiple appendixes use $\backslash${\tt{appendices}} then use $\backslash${\tt{section}} to start each appendix.
% You must declare a $\backslash${\tt{section}} before using any $\backslash${\tt{subsection}} or using $\backslash${\tt{label}} ($\backslash${\tt{appendices}} by itself
%  starts a section numbered zero.)}

\bibliographystyle{IEEEtran}
\bibliography{IEEEabrv, paper}

\vfill

\end{document}

%% file: macros.tex
\usepackage{amsmath,amsfonts}
\usepackage{algorithmic}
\usepackage{algorithm}
\usepackage{array}
\usepackage[caption=false,font=normalsize,labelfont=sf,textfont=sf]{subfig}
\usepackage{textcomp}
\usepackage{stfloats}
\usepackage{url}
\usepackage{verbatim}
\usepackage{graphicx}
\usepackage{cite}
\usepackage[mathscr]{eucal}
\usepackage{balance}
\usepackage{threeparttable}
\usepackage{xcolor}
\usepackage{comment}
\usepackage{booktabs}
\usepackage{multirow}
\usepackage{xspace}
\usepackage{tcolorbox}
\usepackage{adjustbox}
\usepackage{subcaption}
\usepackage{pifont}
\usepackage[normalem]{ulem}

\newcommand{\PreserveBackslash}[1]{\let\temp=\\#1\let\\=\temp}
\newcolumntype{C}[1]{>{\PreserveBackslash\centering}p{#1}}
\newcolumntype{R}[1]{>{\PreserveBackslash\raggedleft}p{#1}}
\newcolumntype{L}[1]{>{\PreserveBackslash\raggedright}p{#1}}

\def\BibTeX{{\rm B\kern-.05em{\sc i\kern-.025em b}\kern-.08em
    T\kern-.1667em\lower.7ex\hbox{E}\kern-.125emX}}

\newcommand{\pa}[1]{\noindent\textbf{#1}}

\newcommand{\toolS}{\space}
\newcommand{\tool}{RobustLOL}
\newcommand{\rpetrans}{\textit{RPE}$_{trans}$}

\newcommand{\rqboxc}[1]{\begin{tcolorbox}[left=4pt,right=4pt,top=4pt,bottom=4pt,colback=gray!5,colframe=gray!40!black,before skip=4pt,after skip=4pt]#1\end{tcolorbox}}

\newcommand{\mathsmall}{%
  \fontsize{8pt}{8pt}\selectfont
}